\documentclass[times,twocolumn,final,authoryear]{elsarticle}

\usepackage{prletters}
\usepackage{framed,multirow}
\usepackage{amsmath,amssymb} 
\usepackage{color}
\usepackage{wrapfig}
\usepackage{booktabs}
\definecolor{grey}{rgb}{0.43, 0.5, 0.5}

\usepackage{amssymb}
\usepackage{latexsym}

\usepackage{url}
\definecolor{newcolor}{rgb}{.8,.349,.1}
\DeclareMathOperator*{\argmin}{arg\,min}

\journal{Pattern Recognition Letters}

\begin{document}

\clearpage

\ifpreprint
  \setcounter{page}{1}
\else
  \setcounter{page}{1}
\fi

\begin{frontmatter}

\title{On the benefits of defining vicinal distributions in latent space}

\author[addr1]{Puneet \snm{Mangla}  \corref{cor1}} 
\ead{cs17btech11029@iith.ac.in, pmangla261@gmail.com}
\author[addr1]{Vedant \snm{Singh} \corref{cor2}}
\ead{cs18btech11047@iith.ac.in}
\author[addr1]{Shreyas \snm{Havaldar} \corref{cor2}}
\ead{cs18btech11042@iith.ac.in}
\author[addr1]{Vineeth \snm{Balasubramanian} \corref{cor3}}
\ead{vineethnb@cse.iith.ac.in}
\cortext[cor1]{Corresponding author}
\address[addr1]{Department of Computer Science and Engineering, IIT Hyderabad, Sangareddy, 502284, India}

\received{1 May 2013}
\finalform{10 May 2013}
\accepted{13 May 2013}
\availableonline{15 May 2013}
\communicated{S. Sarkar}

\begin{abstract}
 The vicinal risk minimization (VRM) principle is an empirical risk minimization (ERM) variant that replaces Dirac masses with vicinal functions. There is strong numerical and theoretical evidence showing that VRM outperforms ERM in terms of generalization if appropriate vicinal functions are chosen. Mixup Training (MT), a popular choice of vicinal distribution, improves generalization performance of models by introducing globally linear behavior in between training examples. Apart from generalization, recent works have shown that mixup trained models are relatively robust to input perturbations/corruptions and at same time are calibrated better than their non-mixup counterparts.
In this work, we investigate the benefits of defining these vicinal distributions like mixup in latent space of generative models rather than in input space itself. We propose a new approach - \textit{VarMixup (Variational Mixup)} - to better sample mixup images by using the latent manifold underlying the data. Our empirical studies on CIFAR-10, CIFAR-100 and Tiny-ImageNet demonstrates  that  models trained by performing mixup in the latent manifold  learned  by  VAEs  are  inherently  more  robust to various input corruptions/perturbations, are significantly better calibrated and exhibit more local-linear loss landscapes.\\
\textbf{\textit{Keywords}} : Robustness $\circ$  Calibration $\circ$ Mixup $\circ$ VRM $\circ$ Common Corruptions
\end{abstract}
\end{frontmatter}

\section{Introduction}
\label{sec_intro}
Deep Neural Networks (DNNs) have become a key ingredient to solve many challenging tasks like classification, segmentation, object detection, speech recognition, etc. In most successful applications, these networks are trained to minimize the average error over the training dataset known as the Empirical Risk Minimization (ERM) principle \citep{Vapnik1998erm}. However, various empirical and theoretical studies have shown that minimizing Empirical Risk over training datasets in over-parameterized settings leads to memorization and thus poor generalization on examples just outside the training distribution. Some classical results in learning theory \citep{vapnik1999vrm} tells us that the convergence of ERM is guaranteed as long as the size of the learning machine (in terms of number of parameters or VC-complexity \citep{harvey2017vcdim}) does not increase with the number of training data. To mitigate this problem of memorization in over-parameterized neural networks, Vicinal Risk Minimization (VRM) was proposed which essentially chooses to train networks on similar but different examples to the training data. This technique more popularly known as data augmentation \citep{Simard2012dataaug}, requires one to define a vicinity or neighbourhood around each training example (eg. in terms of brightness, contrast, imperceptible noise, to name a few). Once defined, more examples can be sampled from their vicinity to enlarge the support of training distribution.

One of the popular choices to create the vicinal distribution is Mixup. Mixup Training (MT) \citep{mixup} has emerged as a popular technique to train models for better generalisation in the last couple of years. Recent works have also shown that the idea of Mixup and Mixup training can be leveraged during inference \citep{mixup-inf} and in many existing techniques like data augmentation \citep{augmix}, adversarial training \citep{IAT}, etc. to improve the robustness of models to various input perturbations \citep{szegedy2013intriguing,adv-exmp-survey} and corruptions \citep{augmix}. Another variant of Mixup, known as as Manifold Mixup \citep{manifold-mixup} encourages neural networks to predict less confidently on interpolations of hidden representations by leveraging semantic interpolations as an additional training signal. As a result, neural networks trained with Manifold Mixup learn class representations with fewer directions of variance.
Other efforts on Mixup \citep{mixup-calibration} have shown that Mixup-trained networks are significantly better calibrated and less prone to over-confident predictions on out-of-distribution than the ones trained in the regular fashion. 


Although still in its early phase, the above efforts \citep{mixup,manifold-mixup,mixup-inf, mixup-calibration} also indicate a trend to view Mixup from perspectives of robustness and calibration.
In this work, we take another step in this direction and propose a new vicinal distribution/sampling technique called \textit{VarMixup (Variational Mixup)} to sample better Mixup images during training to induce robustness as well as improve predictive uncertainty of models while preserving the clean data performance to extent possible. 
In particular, we hypothesize that the latent unfolded manifold underlying the data (through a generative model, a Variational Autoencoder in our case) is linear by construction (manifolds unfold the locally linear structure of a high-dimensional data space), and hence more suitable for the defining vicinal distributions involving linear interpolations, such as Mixup. Importantly, we show that this choice of the distribution for Mixup plays an important role towards robustness and predictive uncertainty (Section \ref{sec_methodology}). We note herein that our downstream task is image classification and the VAE used in our approach is an auxiliary tool rather than being the model to be trained in the first place.

\noindent Our contributions can be summarized as follows:
\begin{itemize}
    \item We propose a new sampling technique called \textit{VarMixup (Variational Mixup)} to sample better Mixup images during training by using the latent manifold learned by generative models. Our experiments on 3 standard datasets- CIFAR-10, CIFAR-100 and Tiny-ImageNet show that VarMixup significantly boosts the robustness to out-of-distribution shifts as well calibration of neural networks as compared to regular mixup or manifold-mixup training. 
    
    \item We conduct additional analysis/studies which show that VarMixup significantly decreases the local linearity error of the neural network and generates samples that are slightly off-distribution from training examples or mixup generated samples, to provide robustness.
\end{itemize}

\section{Background and Related Work}
\label{sec_related_work}
\subsection{Notations and Preliminaries}
We denote a neural network as $F_w : \mathbb{R}^{c\times h \times w} \rightarrow \mathbb{R}^k$, with weight parameters $w$. 
$F_w$ takes an image $x \in  \mathbb{R}^{c\times h \times w}$ and outputs logits, $F^i_w(x)$ for each class $i \in \{1...k\}$. Without loss of generality, we assume the classification task with $\mathcal{L}$ as the standard cross-entropy loss function.  $p_{actual}$ denotes the training data distribution, and the optimal weight parameter $w^*$ is obtained by training the network using standard empirical risk minimization \citep{Vapnik1998erm}, i.e. $w^* = \argmin_w \mathbb{E}_{(\textbf{x},y) \sim p_{actual}} \ \left[ \ \mathcal{L} \ (F_w(\textbf{x}),y) \ \right] $, where $y$ is the true label associated with input $x$.

\subsection{Vicinal Risk Minimization}
Given the data distribution $p_{actual}$, a neural network $F_w$ and loss function $\mathcal{L}$, the \textit{expected risk} (average of loss function over $p_{actual}$) is given by $R(F_w) = \int \mathcal{L}(F_w(x),y) \cdot dp_{actual}(x,y)$.
In practice, the true distribution $p_{actual}$ is unknown, and is approximated by the training dataset $D = \{(x_i,y_i)\}_{i=1}^N$, which represents the \textit{empirical distribution}: $p_\delta(x,y) = \frac{1}{N} \cdot \sum_{i=1}^N \delta(x=x_i, y=y_i)$.
Here, $\delta(x=x_i, y=y_i)$ is the Dirac delta function centered at $(x_i,y_i)$. Using $p_\delta$ as an estimate to $p_{actual}$, we define \textit{expected empirical risk} as:
\begin{equation}
\label{erm-eq}
    R_\delta(F_w) = \frac{1}{N} \cdot \sum_{i=1}^N \mathcal{L}(F_w(x_i),y_i)
\end{equation}
Minimizing Eqn \ref{erm-eq} to find optimal $F_{w^*}$ is typically termed \textit{Empirical Risk Minimization (ERM)} \citep{Vapnik1998erm}. However overparametrized neural networks can suffer from memorizing, leading to undesirable behavior of network outside the training distribution, $p_\delta$ \citep{zhang2017memorize, szegedy2013intriguing}. Addressing this concern, \citep{vapnik1999vrm} and \citep{chapelle2001vrm} proposed \textit{Vicinal Risk Minimization (VRM)}, where $p_{actual}$ is approximated by a vicinal distribution $p_v$, given by:
\vspace{-6pt}
\begin{equation}
    p_v(x,y) = \frac{1}{N} \cdot \sum_{i=1}^N v(x,y \vert x_i, y_i)
\vspace{-6pt}
\end{equation}
\noindent where $v$ is the \textit{vicinal distribution} that calculates the probability of a data point $(x,y)$ in the vicinity of other samples $(x_i,y_i)$. Thus, using $p_v$ to approximate $p_{actual}$, \textit{expected vicinal risk} is given by:
\vspace{-6pt}
\begin{equation}
    R_v(F_w) = \frac{1}{N} \cdot \sum_{i=1}^N g(F_w,\mathcal{L},x_i,y_i)
\vspace{-6pt}
\end{equation}
where $g(F_w,\mathcal{L},x_i,y_i) = \int \mathcal{L}(F_w(x),y) \cdot dv(x,y \vert x_i,y_i)$. The superiority of VRM over ERM has been theoretically as well as empirically verified by many recent works \citep{Ni2015vrm,cao2015vrm,Luan2010vrm,zhang2018generalization}.

Popular examples of vicinal distributions include: (i) \textit{Gaussian Vicinal distribution}: Here, $v_{gaussian}(x,y \vert x_i, y_i) = \mathcal{N}(x-x_i, \sigma^2) \cdot \delta(y=y_i)$, which is equivalent to augmenting the training samples with Gaussian noise; and (ii) \textit{Mixup Vicinal distribution} : Here $v_{mixup}(x,y \vert x_i, y_i) = \frac{1}{n} \cdot \sum_{j=1}^N \mathbb{E}_\lambda [ \delta(x = \lambda \cdot x_i + (1-\lambda) \cdot x_j, y = \lambda \cdot y_i + (1-\lambda) \cdot y_j) ]$, where $\lambda \sim  \beta(\eta,\eta)$ and $\eta > 0$.

\subsection{Mixup}
\citep{mixup} proposed Mixup, a method to train models on the convex combination of pairs of examples and their labels. In other words, it constructs virtual training examples as: $x' = \lambda \cdot x_i + (1 - \lambda) \cdot x_j; 
y' = \lambda \cdot y_i + (1 - \lambda) \cdot y_j$, where $x_i$, $x_j$  are input vectors; $y_i$, $y_j$ are one-hot label encodings and $\lambda$ is a mixup coefficient, usually sampled from a $\beta(\eta,\eta)$ distribution. By doing so, it regularizes the network to behave linearly in between training examples, thus inducing global linearity between them. A recent variant, Manifold Mixup \citep{manifold-mixup}, exploits interpolations at hidden representations, thereby obtaining neural networks with smoother decision boundaries at different levels of hidden representations. AugMix \citep{augmix} mixes up multiple augmented images and uses a Jensen-Shannon Divergence consistency loss on them to achieve better robustness to common input corruptions \citep{hendrycks-robustness}. In semi-supervised learning, MixMatch \citep{mixmatch} obtains state-of-the-art results by guessing low-entropy labels for data-augmented unlabeled examples and mixes labeled and unlabeled data using Mixup. It has been shown that apart from better generalization, Mixup also improves the robustness of models to adversarial perturbations as well. To further boost this robustness at inference time, Pang et al. \citep{mixup-inf} recently proposed a Mixup Inference technique which performs a mixup of input $x$ with a clean sample $x_s$ and passes the corresponding mixup sample ($\lambda \cdot x + (1-\lambda) \cdot x_s$) into the classifier as the processed input. 
Other efforts related to Mixup \citep{mixup-calibration} have shown that Mixup-trained networks are better calibrated i.e., the predicted softmax scores are better indicators of the actual likelihood of a correct prediction than DNNs trained in the regular fashion. Additionally, they also observed that mixup-trained DNNs are less prone to over-confident predictions on out-of-distribution and random-noise data. None of these efforts however address Mixup from a generative latent space, which is the focus of this work. Efforts such as \citep{mixup-inf} and \citep{mixup-calibration}, in fact, have inferences that motivate the need to consider a latent Mixup space to address a model's robustness and predictive uncertainty.
\vspace{-1pt}
From a different perspective, Xu et al. \citep{domain-mixup} used domain mixup to improve the generalization ability of models in domain adaptation. Adversarial Mixup Resynthesis \citep{adversarial-mixup} attempted mixing latent codes used by autoencoders through an arbitrary mixing mechanism that can recombine codes from different inputs to produce novel examples. This work however has a different objective and focuses on generative models in a Generative Adversarial Network (GAN)-like setting, while our work focuses on robustness and predictive uncertainty. The work by Liu et al. \citep{latent-mix} may be closest to ours in terms of approach as they use an adversarial autoencoder (AAE) to impose a uniform distribution on the feature representations. However, their work deals with improving generalization performance, while ours looks at robustness and predictive uncertainty, as already stated. Other related works like \citep{gan_robust1, gan_robust2, gan_robust3} attempt to leverage Generative Adversarial Networks (GANs) to train adversarially robust classifiers, while, we focus on VAEs, because of their ability to model the latent manifold  explicitly, to train classifiers robust to commonly observed out-of-distribution shifts (eg. snow, fog etc.)  (explained in detail in Section \ref{sec_methodology}). Furthermore, we propose a new method, VarMixup, which focuses on directly exploiting the manifold learned by a Variational Autoencoder (VAE) (and do not regularize it unlike previous work) during Mixup and report improved adversarial robustness. We also present useful insights into the working of VarMixup (which is lacking in earlier work including \citep{latent-mix}), thus making our contributions unique and more complete.
\section{Methodology}
\label{sec_methodology}
In this work, we build on the recent success of using Mixup as a vicinal distribution by proposing the use of the latent spaces learned by a generative deep neural network model.
The use of generative models such as Variational Autoencoders (VAEs) \citep{VAE} to capture the latent space from which a distribution is generated provides us an unfolded manifold (the low-dimensional latent space), where the linearity in between training examples is more readily observed. Defining vicinal distributions by using neighbors on this latent manifold, which is more linear in the low-dimensional space, learned by generative models provides us more effective linear interpolations than the ones in input space. We hence leverage such an approach to capture the induced global linearity in between examples, and define Mixup vicinal distributions on this latent surface.

\subsection{Our Approach: VarMixup (Variational Mixup)}
\begin{figure*}
    \centering
    \includegraphics[width=0.7\linewidth]{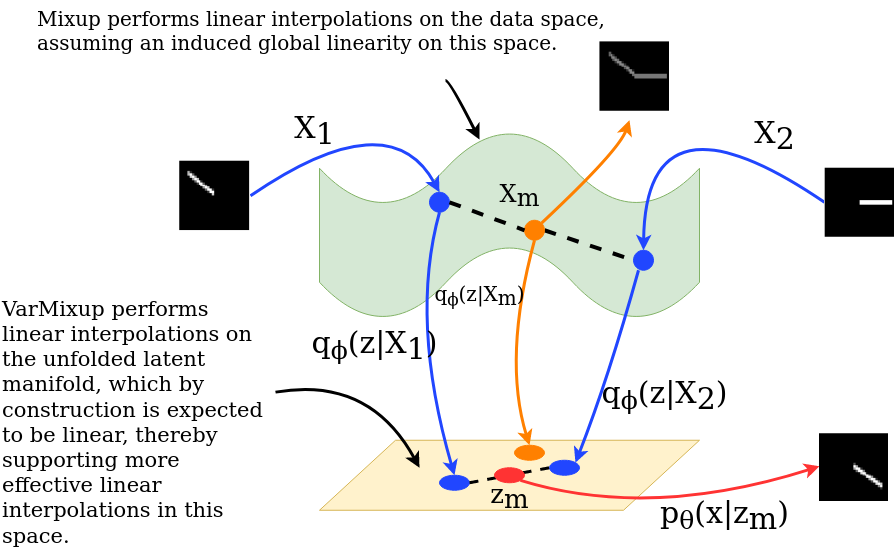}
    \caption{\footnotesize Illustration of the conceptual idea behind VarMixup. We interpolate on the unfolded manifold, as defined by a generative model (VAE, in our case).}

    \label{fig:architecture}
\end{figure*}

To capture the latent manifold of the training data through a generative model, we opt for a Variational Autoencoder (VAE).  VAE \citep{VAE} is an autoencoder which is trained using Variational Inference, which serves as an implicit regularizer to ensure that the obtained latent space allows us to generate new data from the same distribution as training data. Our rationale behind choosing VAEs over GANs to capture latent manifold of training data is that while VAEs are known for modeling latent variable models explicitly, GANs are implicit generative models, i.e. while we can generate images from latent variables, the reverse operation - getting latent variable samples corresponding to images - is not explicitly modeled. To obtain the latent embedding of an image $x$, one may have to solve the following optimization via backpropogation:
\begin{equation}
    z^* = \argmin_z \Vert G(z) - x \Vert_2
\end{equation}
where $G$ is the generator. Clearly, such an optimization causes additional overhead in increased time and computation complexity. Since we work on defining vicinal distributions in the latent space, choosing VAEs directly allows us to obtain the latent embeddings via one forward pass. \par
We denote the encoding and decoding distribution of VAE as $q_\phi(z \vert x)$ and $p_\theta(x \vert z)$ respectively, parametrized by $\phi$ and $\theta$ respectively. Given $p(z)$ as the desired prior distribution for encoding, the general VAE objective is given by the loss function:
\begin{equation} 
\label{vae-obj}
    \mathcal{L}_{VAE} = -\gamma \cdot D(q_\phi(z) \Vert p(z)) + \mathbb{E}_{x \sim p_{actual}} \mathbb{E}_{z \sim q_\phi(z \vert x)} [\log(p_\theta(x \vert z))]
\end{equation}
Here, $D$ is any strict divergence, meaning that $D(q \Vert p) \geq 0$ and $D(q \Vert p) = 0$ if and only if $q=p$, and $\gamma>0$ is a scaling coefficient. The second term in the objective acts as a image reconstruction loss and $q_\phi(z) = \mathbb{E}_{x \sim p_{actual}} [q_\phi(z \vert x)]$.
The original VAE \citep{VAE} uses KL-divergence in Eqn \ref{vae-obj}, and thus optimizes the following objective: 
\begin{equation}
\label{vanilla-vae-obj}
    \mathcal{L}_{VAE} =  \mathbb{E}_{x \sim p_{actual}} \left[ -\gamma KL(q_\phi(z|x) \Vert p(z))] + \mathbb{E}_{z \sim q_\phi(z \vert x)} \log(p_\theta(x \vert z)) \right ]
\end{equation}
\noindent However, using KL-divergence in Eqn \ref{vae-obj} has some shortcomings, as pointed out in \citep{vae-drawback2,vae-drawback1,MMD-VAE, LadderVAE}. 
KL-divergence encourages the encoding $q_\phi(z|x)$ to be a random sample from $p(z)$ for each $x$, making them uninformative about the input. Also, it is not strong enough a regularizer compared to the reconstruction loss and tends to overfit data, consequently, learning a $q_\phi(z|x)$ that has high variance. Both the aforementioned shortcomings can affect the encoding distribution by making them uninformative of inputs with high variance. Since we use VAEs to better capture a linear latent manifold and subsequently define interpolations there, a bad latent distribution can affect our method significantly. Hence, we use a variant \textit{Maximum Mean Discrepancy VAE} (MMD-VAE)\citep{MMD-VAE} which uses a MMD Loss \citep{MMD} instead of KL-divergence, and hence optimizes the following objective:
\begin{equation} 
\label{mmd-vae-obj}
    \mathcal{L}_{MMD-VAE} = \gamma \cdot MMD(q_\phi(z) \Vert p(z)) + \mathbb{E}_{x \sim p_{actual}} \mathbb{E}_{z \sim q_\phi(z \vert x)} [\log(p_\theta(x \vert z))]
\end{equation}
A MMD-VAE doesn't suffer from the aforementioned shortcomings \citep{MMD-VAE}, as it maximizes mutual information between $x$ and $z$ by matching the distribution over encodings $q_\phi(z)$ with prior $p(z)$ only in expectation, rather than for every input. We hence train an MMD-VAE to characterize the training distribution more effectively.
Once trained, we now define a Mixup vicinal distribution in the latent space of the trained VAE as:
\begin{equation}
\label{varmix}
\begin{split}
    v_{VarMixup}(z,y \vert x_i, y_i) = & \frac{1}{n} \cdot \sum_{j=1}^N \mathbb{E}_\lambda [ \delta(z = \lambda \cdot \mathbb{E}_z [q_\phi(z \vert x_i)] + 
    \\ & (1-\lambda) \cdot \mathbb{E}_z [q_\phi(z \vert x_j)], 
    \vspace{-2pt} \\
    &  y = \lambda \cdot y_i + (1-\lambda) \cdot y_j) ]
\end{split}
\end{equation}
where $\lambda \sim  \beta(\eta,\eta)$ and $\eta > 0$. Using the above vicinal distribution, $v_{VarMixup}$ and the MMD-VAE decoder, $p_\theta(x \vert z)$, we construct VarMixup samples as:
\vspace{-4pt}
\begin{equation}
\label{varmixup_sample}
    \begin{split}
    x' & = \mathbb{E}_x [p_\theta(x \vert \lambda \cdot \mathbb{E}_z [q_\phi(z \vert x_i)] + (1 - \lambda) \cdot \mathbb{E}_z [q_\phi(z \vert x_j)])]\\ \vspace{-2pt} 
    y' & = \lambda \cdot y_i + (1 - \lambda) \cdot y_j
    \vspace{-4pt}
    \end{split}
\end{equation}
From another perspective, one could view our new sampling technique as performing Manifold Mixup \citep{manifold-mixup}, however over the latent space of an MMD-VAE (instead of the neural network feature space) and using it for sample reconstruction. We compare against Manifold Mixup in our results to show the improved performance of the learned generative latent space in our VarMixup. Figure \ref{fig:architecture} illustrates the conceptual idea behind VarMixup. The entire training methodology of VarMixup can be summarized in the following steps:
\begin{itemize}
\setlength\itemsep{-0.1em}
    \item Train an MMD-VAE using Equation \ref{mmd-vae-obj}.
    \item Generate VarMixup samples $\{(x^{(i)}, y^{(i)})\}$ according to Equation \ref{varmixup_sample}.
    \item Optimize model on generated VarMixup samples, $\{(x^{(i)}, y^{(i)})\}$ via standard cross-entropy loss.
\end{itemize}


\section{Experiments and Results}
\label{sec_expts}
We now present our experimental studies and results using our method, VarMixup, on multiple datasets. We begin by describing the datasets, evaluation criteria and implementation details. Note that we focus explicitly on the usefulness of our approach on out-of-distribution test data and addressing predictive uncertainty\\
\noindent \textbf{Datasets:} We perform experiments on three well-known standard datasets: CIFAR-10, CIFAR-100 \citep{cifar-datasets} and Tiny-ImageNet \citep{tiny-img}. \textit{CIFAR-10} is a subset of 80 million tiny images dataset and consists of 60,000 $32 \times 32$ color images containing one of 10 object classes, with 6000 images per class. \textit{CIFAR-100} is just like CIFAR-10, except that it has 100 classes containing 600 images each. There are 500 training images and 100 testing images per class. \textit{Tiny-Imagenet} has 200 classes, with each class containing 500 training images, 50 validation images, and 50 test images. Each image here is of resolution $64 \times 64$.

\noindent \textbf{Evaluation Criteria:} 
To measure the generalization of our models on out-of-distribution data, we evaluate their robustness on the newer CIFAR-10-C, CIFAR-100-C and Tiny-Imagenet-C datasets \citep{hendrycks-robustness}. These datasets contain images, corrupted with 15 different distortions at 5 severity levels (Gaussian blur, Shot Noise, Impulse Noise, JPEG compression, Motion blur, frost, to name a few). For completeness, we also report accuracy on clean images and standard deviations over 10 trials (which captures standard generalization performance). We also measure the Expected Calibration Error (ECE) \citep{calibration} of our trained models to quantify their predictive uncertainty. 

\noindent \textbf{Implementation Details:}
It has been shown \citep{adv-feature} that adversarial robust training \citep{at@madry} removes irrelevant biases (e.g. texture biases) in their hidden representations, thus making them more informative. We hence hypothesize that the considered VAE, if trained in an adversarially robust fashion, will have more informative latent encoding than its regular equivalent. This would hence help improve the empirical/vicinal distributions like VarMixup. Empirically, we validate this hypothesis in our subsequent experiments and use prefix \textit{adv-} (eg: \textit{adv}-VarMixup) to distinguish them from their regular variants. We note that the aforementioned approach of adversarial robust training \citep{at@madry} is different from adversarial training used to train GAN-like architectures, and hence both should not be confused. In other words, the adversarial robust training that we are referring to, minimizes adversarial ELBO instead of standard ELBO (\ref{mmd-vae-obj}) of an MMD-VAE. Given a dataset $\mathcal{D} = \{ (x_i, y_i) \}$, we identify adversarial ELBO $\mathcal{L}_{MMD-VAE}^{adv}$ as follows: 

\begin{equation}
    \begin{split}
    \because q_\phi(z) & = \mathbb{E}_{x \sim p_{actual}} [q_\phi(z \vert x)] = \frac{1}{\vert \mathcal{D} \vert} \sum_i q_\phi(z \vert x_i) \\
        \mathcal{L}_{MMD-VAE} \ (x_1, ... x_i) & = \gamma \cdot MMD \left (\frac{1}{\vert \mathcal{D} \vert} \sum_i q_\phi(z \vert x_i) \ \Vert \ p(z)\right ) \\
        & + \frac{1}{\vert \mathcal{D} \vert} \sum_i \mathbb{E}_{z \sim q_\phi(z \vert x_i)} \ [ \ \log(p_\theta(x_i \ \vert \ z))\ ] \\
        x_i^* & = \max_{x_i \ \in \ \mathcal{B}(x_i, \epsilon)} \mathcal{L}_{MMD-VAE}\ (x_1, ... x_i) \\ \mathcal{L}_{MMD-VAE}^{adv} \ (x_1, ... x_i) & = \gamma \cdot MMD \left (\frac{1}{\vert \mathcal{D} \vert} \sum_i q_\phi(z \vert x^*_i) \ \Vert \ p(z)\right ) \\
        & + \frac{1}{\vert \mathcal{D} \vert} \sum_i \mathbb{E}_{z \sim q_\phi(z \vert x^*_i)} \ [ \ \log(p_\theta(x^*_i \ \vert \ z))\ ]
    \end{split}
\end{equation}

We choose Resnet-34 \citep{resnet} and WideResNet-28-10 \citep{Zagoruyko2016WRN} (SOTA backbone \cite{mixup, manifold-mixup, at@madry}) as the backbone architecture for evaluating our approach and baselines. 

\vspace{3pt}
\noindent \textbf{Baseline Models:} We compare our method, VarMixup, against an exhaustive set of baselines including non-VRM variants, mixup variants and state-of-the-art adversarial techniques. Below are their details:
\begin{enumerate}
  \item \textit{ERM} - Vanilla Empirical Risk Minimization (Eqn \ref{erm-eq}) using Adam optimizer ($lr=1e-3$) for 100 epochs on all datasets.
  \item \textit{Mixup} - Vanilla Mixup training \citep{mixup} using Adam optimizer ($lr=1e-3$) for 150 epochs on all datasets. Mixup coefficient is sampled from $\beta(1,1)$.
  \item \textit{Mixup-R} - Mixup training on MMD-VAE's reconstructed image space \citep{mixup} using Adam optimizer ($lr=1e-3$) for 150 epochs on all datasets. Mixup coefficient is sampled from $\beta(1,1)$.
  \item \textit{Manifold Mixup} - Manifold Mixup training \citep{manifold-mixup} using Adam optimizer ($lr=1e-3$) for 150 epochs on all datasets. Mixup coefficient is sampled from $\beta(2,2)$.
  \item \textit{AT and TRADES}  - $l_\infty$ PGD/TRADES adversarial training \citep{at@madry,trades} with $\epsilon = 8/255$ and step-size $\alpha=2/255$. Models are trained using Adam optimizer ($lr=1e-3$) for 250 epochs on all datasets.
  \item \textit{IAT}  - $l_\infty$ Interpolated adversarial training \citep{IAT} with $\epsilon = 8/255$ and step-size $\alpha=2/255$. Interpolation coefficient is sampled from $\beta(1,1)$. Models are trained using Adam optimizer ($lr=1e-3$) for 350 epochs on all datasets.
\end{enumerate}

\noindent \textbf{Generalization Performance and Robustness to Out-of-Distribution shifts}
We first evaluate the trained models on their robustness to various common input corruptions, along with their generalization performance on ``clean data'' (test data without corruptions). Hendrycks et al \citep{hendrycks-robustness} recently proposed the CIFAR-10-C, CIFAR-100-C, and Tiny-Imagenet-C datasets, which are extensions of CIFAR-10, CIFAR-100 and Tiny-Imagenet containing images corrupted with 15 different distortions and 5 levels of severity. We report the mean classification accuracy over all distortions on these datasets in Table \ref{table:corrupt} and Table \ref{table:corrupt_wrn} for ResNet-34 \cite{resnet} and WideResNet-28-10 \citep{Zagoruyko2016WRN} architectures respectively. The results show that our method - VarMixup/\textit{adv}-VarMixup achieves superior performance by a margin of $\sim 2-10 \%$ consistently across the datasets. 
We observe a slight drop in the clean accuracy of VarMixup models (shown in parentheses in Table \ref{table:corrupt} and Table \ref{table:corrupt_wrn}) which we believe is due to the tradeoff between robustness and clean accuracy, a common trend observed in robustness literature \citep{robustness-odds}. However, in an attempt to strike a balance between both (i.e reduce the tradeoff), we conduct an additional experiment where  we exploit the benefits that Mixup or VarMixup offer individually. More specifically in each training iteration, we randomly choose (with probability of 0.5) to sample either using Mixup or VarMixup distribution. We refer this experiment as Mixup + VarMixup in Table \ref{table:corrupt} and Table \ref{table:corrupt_wrn} and observe that it leads to clean test accuracy comparable to regular Mixup/Manifold-Mixup whilst improving or atleast maintaining similar performance on corrupted benchmarks. \par
Moreover, we would also like to point out that this trade-off between robust and clean accuracies has been a subject of research itself, where efforts like \citep{IAT, avmixup} have proposed methodologies to narrow the gap between the two accuracies. In this work, our primary goal is to improve the robustness of neural networks while preserving performance on clean data to the extent possible. Further study on reducing the trade-off between robust-clean accuracy in this setting is left as a direction of future work.
\begin{table*}
\caption{\footnotesize Robustness to common input corruptions on CIFAR-10-C, CIFAR-100-C and Tiny-Imagenet-C \citep{hendrycks-robustness} datasets using ResNet-34 \cite{resnet} backbone. Best results in \textbf{bold} and second best \underline{underlined}. Clean accuracy is reported in parentheses using gray colour.}
\label{table:corrupt}
\begin{center}
\scalebox{1}{
\begin{tabular}{c|c|c|c}
\toprule
\textbf{Method} & \textbf{CIFAR-10-C} & \textbf{CIFAR-100-C} & \textbf{Tiny-Imagenet-C} \\
\midrule

AT \citep{at@madry} & 73.12 $\pm$ 0.31 \textcolor{grey}{(85.58 $\pm$ 0.14)} & 45.09 $\pm$ 0.31 \textcolor{grey}{(60.28 $\pm$ 0.13)} & 15.74 $\pm$ 0.36 \textcolor{grey}{(22.33 $\pm$ 0.16)}\\
TRADES \citep{at@madry} & 75.46 $\pm$ 0.21 \textcolor{grey}{(88.11 $\pm$ 0.43)} & 45.98 $\pm$ 0.41 \textcolor{grey}{(63.3 $\pm$ 0.32)} & 16.20 $\pm$ 0.23 \textcolor{grey}{(26.12 $\pm$ 0.38)}\\
IAT \citep{IAT} & 81.05	$\pm$ 0.42 \textcolor{grey}{(89.7 $\pm$ 0.33)} & 50.71 $\pm$ 0.25 \textcolor{grey}{(62.7 $\pm$ 0.21)} & 18.69 $\pm$ 0.45 \textcolor{grey}{(18.08 $\pm$ 0.34 )}\\
\midrule

ERM	& 69.29	$\pm$ 0.21 \textcolor{grey}{(94.5 $\pm$ 0.14)} & 47.3 $\pm$ 0.32 \textcolor{grey}{(64.5 $\pm$ 0.10)} & 17.34 $\pm$ 0.27 \textcolor{grey}{(49.96 $\pm$ 0.12)}\\
Mixup & 74.74	$\pm$ 0.34 \textcolor{grey}{(95.5 $\pm$ 0.35 )} & 52.13 $\pm$ 0.43 \textcolor{grey}{(76.8 $\pm$ 0.41)} & 21.55 $\pm$ 0.37 \textcolor{grey}{(53.83 $\pm$ 0.17)}\\
Mixup-R & 74.27 $\pm$ 0.22 \textcolor{grey}{(89.88 $\pm$ 0.11)} & 43.54 $\pm$ 0.15 \textcolor{grey}{(62.24 $\pm$ 0.21)} & 21.34 $\pm$ 0.32 \textcolor{grey}{(53.5 $\pm$ 0.28)}\\
Manifold-Mixup & 72.54 $\pm$ 0.14 \textcolor{grey}{(95.2 $\pm$ 0.18)} & 41.42 $\pm$ 0.23 \textcolor{grey}{(75.3 $\pm$ 0.48)} & -\\
VarMixup & \underline{82.57 $\pm$ 0.42} \textcolor{grey}{(93.91 $\pm$ 0.45 )}	& 52.57 $\pm$ 0.39 \textcolor{grey}{(73.2 $\pm$ 0.44 )} & 24.87 $\pm$ 0.32 \textcolor{grey}{(50.98 $\pm$ 0.11)}\\
\textit{adv}-VarMixup & 82.12 $\pm$ 0.46 \textcolor{grey}{(92.19 $\pm$ 0.32 )}	& \underline{54.0 $\pm$ 0.41} \textcolor{grey}{(72.13 $\pm$ 0.34)} & \underline{25.36 $\pm$ 0.21} \textcolor{grey}{(50.58 $\pm$ 0.23)} \\
Mixup + VarMixup & \textbf{83.36 $\pm$ 0.46} \textcolor{grey}{(94.1 $\pm$ 0.13 )}	& \textbf{54.36 $\pm$ 0.05} \textcolor{grey}{(75.3 $\pm$ 0.23)} & \textbf{26.87 $\pm$ 0.21} \textcolor{grey}{(52.58 $\pm$ 0.47)} \\
\bottomrule
\end{tabular}}

\end{center}
\end{table*} 


\vspace{3pt}
\noindent \textbf{Calibration:}
A recent study \citep{mixup-calibration} showed that DNNs trained with Mixup are significantly better calibrated than DNNs trained in a regular fashion. Calibration \citep{calibration} measures how good softmax scores are as indicators of the actual likelihood of a correct prediction. We measure the \textit{Expected Calibration Error (ECE)} \citep{mixup-calibration,calibration} of the proposed method, following \citep{mixup-calibration}: predictions (total $N$ predictions) are grouped into $M$ interval bins ($B_m$) of equal size. The accuracy and confidence of $B_m$ are defined as: 
\[acc(B_m) = \frac{1}{\vert B_m \vert} \sum_{i \in B_m} 1 \cdot (\hat{y_i}= y_i)\] 
and 
\[conf(B_m) = \frac{1}{\vert B_m \vert} \sum_{i \in B_m} \hat{p_i}\] 
where $\hat{p_i}$, $\hat{y_i}$, $y_i$ are the confidence, predicted label and true label of sample $i$ respectively. The \textit{Expected Calibration Error (ECE)} is then defined as: 
\begin{equation}
    ECE = \sum_{m=1}^M \frac{\vert B_m \vert}{N} \cdot \vert \ acc(B_m) - conf(B_m) \ \vert \
\end{equation}

\begin{figure}
\centering
    \includegraphics[width=\linewidth]{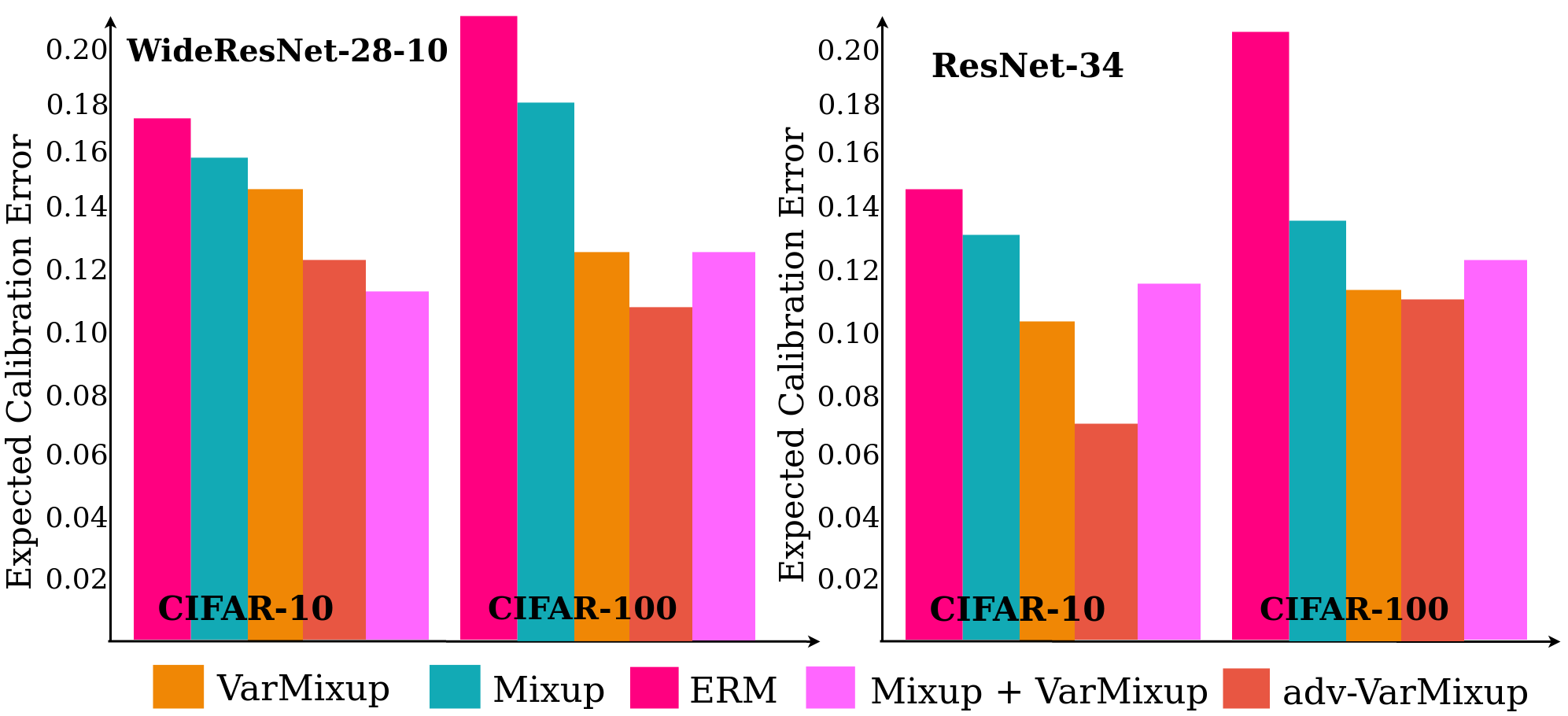}
    \caption{\footnotesize Expected Calibration Error (ECE) \citep{calibration} of ERM, Mixup, VarMixup, adv-VarMixup, Mixup + VarMixup trained models.}
    \label{fig:calibration}
\end{figure}

\noindent Figure \ref{fig:calibration} shows the calibration error on CIFAR-10 and CIFAR-100 datasets using Mixup, VarMixup, adv-VarMixup and Mixup + VarMixup. The figure illustrates that our VarMixup models (and their combinations with regular Mixup) are also better calibrated than regular Mixup. 
\section{Discussion and Ablations}
\label{sec_ablations}
\noindent \textbf{Local linearity on loss landscapes:}
\citep{local-lin} showed that the local linearity of loss landscapes of neural networks is related to model robustness. The more the loss landscapes are linear, the more the adversarial robustness. To further study this observation using our method, we analyze the local linearity of loss landscapes of VarMixup and regular mixup trained models. Qin et al. \citep{local-lin} defines local linearity at a data-point $x$  within a neighbourhood $B(\epsilon)$ as $\gamma(\epsilon,x,y) = $
\begin{equation}
    \max_{\delta \in B(\epsilon)} \vert \mathcal{L}(F_w(x + \delta),y) - \mathcal{L}(F_w(x ),y) - \delta^T  \triangledown_x  \mathcal{L}(F_w(x ),y) \vert
\end{equation}

\begin{figure}
\centering
    \includegraphics[width=0.9\linewidth]{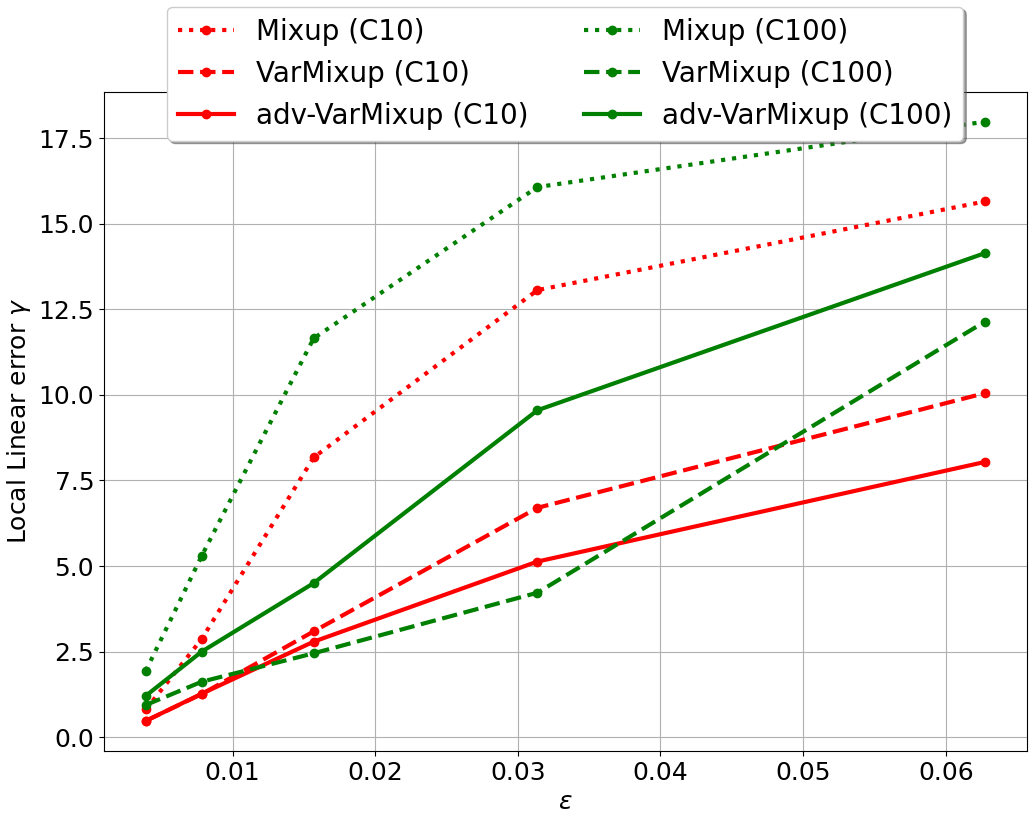}
    \caption{\footnotesize Local linear error of loss landscapes of the models trained on CIFAR-10/-100 (denoted as C10 and C100)}
    \label{fig:lle}
\end{figure}

Figure \ref{fig:lle} shows the average local linear error (over test set) with increasing $L_\infty$ max-perturbation $\epsilon$ on CIFAR-10 and CIFAR-100 datasets. As noticeable, VarMixup/\textit{adv}-VarMixup makes the local linear error significantly ($\times$ 2) lesser as compared to regular mixup, thus inducing robustness. 

\begin{figure}
    \centering
    \includegraphics[width=\linewidth]{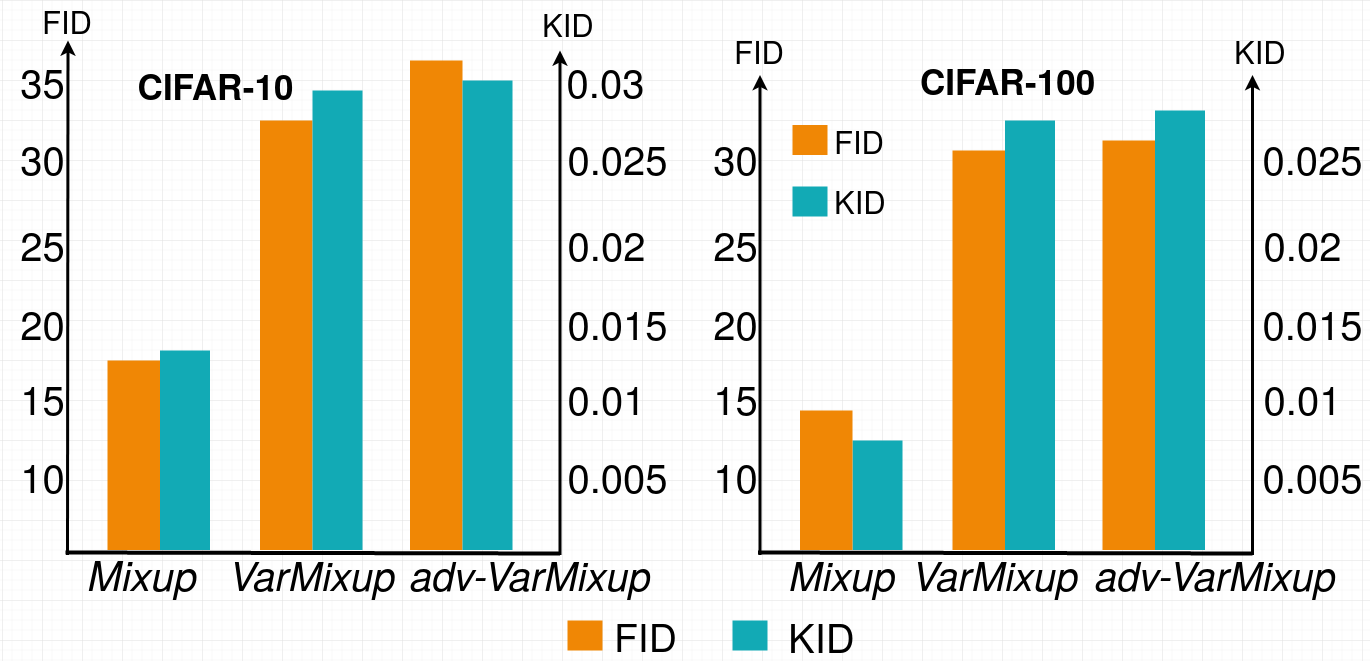}
    \caption{\footnotesize FID and KID scores between training set and Mixup/VarMixup generated samples on CIFAR-10 and CIFAR-100}
    \label{fig:fid}
\end{figure}

\begin{figure}[h]
    \centering
    \includegraphics[width=0.8\linewidth]{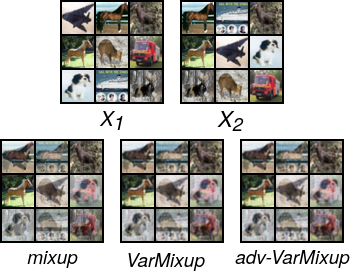}
    \caption{\footnotesize Samples generated by mixup, VarMixup and \textit{adv}-VarMixup on CIFAR-10 (Mixup coefficient $\lambda=0.5$).}
    \label{fig:sample2}
\end{figure}

\vspace{3pt}
\noindent \textbf{Analyzing VarMixup samples:}
Figure \ref{fig:sample2} shows sample data generated by regular Mixup, VarMixup, and \textit{adv}-VarMixup on two images. Although mixup or VarMixup samples look perceptually similar, they are quite different at a statistical level.
We measure the Frechet Inception Distance (FID) \citep{FID} and Kernel Inception Distance \citep{KID} between regular training data and training data generated by mixup/VarMixup/ \textit{adv}-VarMixup. These scores summarize how similar the two groups are in terms of statistics on computer vision features of the raw images calculated using the Inceptionv3 model used for image classification. 
Lower scores indicate the two groups of images are more similar, or have more similar statistics, with a perfect score being 0.0 indicating that the two groups of images are identical. Figure \ref{fig:fid} reports these metrics on CIFAR-10 and CIFAR-100 respectively. The greater FID and KID scores indicate that we are adding off-manifold samples (w.r.t. the manifold characterized by training data) to the training using our approach.

\vspace{3pt}
\noindent \textbf{Computational Overhead:}
We compare the computational time of our trained models using VarMixup/\textit{adv}-VarMixup with commonly used adversarial training techniques: AT and TRADES. VarMixup, \textit{adv}-VarMixup, AT and TRADES take around 3, 5, 8.8 and 15 hours respectively for training. The training time of the MMD-VAE was also considered here. While already significantly faster than AT and TRADES, the proposed method will be more scalable and time-efficient, if a VAE trained on a dataset such as ImageNet can be directly used to generate VarMixup samples for other datasets. This is a typical transfer learning setting, and we hence study the performance of training VarMixup models on CIFAR-10 and CIFAR-100 datasets using MMD-VAE trained on the Tiny-Imagenet dataset. Respectively, VarMixup obtained mean corruption accuracy of 82.0 \% and 53.25 \% on CIFAR-10-C and CIFAR-100-C benchmarks, thus making our approach time-efficient and also scalable.

\begin{table}
\caption{\footnotesize Robustness to common input corruptions on CIFAR-10-C and CIFAR-100-C \citep{hendrycks-robustness} datasets using WideResNet-28-10 \citep{Zagoruyko2016WRN} backbone. Best results in \textbf{bold} and second best \underline{underlined}. Clean accuracy is reported in parentheses using gray colour.}
\label{table:corrupt_wrn}
\begin{center}
\scalebox{0.75}{
\begin{tabular}{c|c|c}
\toprule
\textbf{Method} & \textbf{CIFAR-10-C} & \textbf{CIFAR-100-C} \\
\midrule

AT& 74.8 $\pm$ 0.34 \textcolor{grey}{(87.32 $\pm$ 0.11)} & 46.1 $\pm$ 0.06 \textcolor{grey}{(62.5 $\pm$ 0.20)}\\
TRADES & 77.39 $\pm$ 0.21 \textcolor{grey}{(89.97 $\pm$ 0.53)} & 46.7 $\pm$ 0.22 \textcolor{grey}{(65.6 $\pm$ 0.33)} \\
IAT& 82.25 $\pm$ 0.44 \textcolor{grey}{(91.3 $\pm$ 0.09)} & 52.3 $\pm$ 0.56 \textcolor{grey}{(63.67 $\pm$ 0.77)} \\
\midrule
ERM	& 72.46 $\pm$ 0.12 \textcolor{grey}{(96.0 $\pm$ 0.43)} & 46.7 $\pm$ 0.22 \textcolor{grey}{(77.27 $\pm$ 0.39)}\\
Mixup & 75.62 $\pm$ 0.16 \textcolor{grey}{(97.1 $\pm$ 0.51)} & 52.46 $\pm$ 0.11 \textcolor{grey}{(80.53 $\pm$ 0.37)} \\
Manifold-Mixup & 73.78 $\pm$ 0.31 \textcolor{grey}{(97.3 $\pm$ 0.08)} & 45.6 $\pm$ 0.33 \textcolor{grey}{(81.2 $\pm$ 0.26)} \\
VarMixup & \underline{84.39 $\pm$ 0.22} \textcolor{grey}{(95.81 $\pm$ 0.13)} & 53.78 $\pm$ 0.42 \textcolor{grey}{(77.24 $\pm$ 0.56)} \\
\textit{adv}-VarMixup & \textbf{84.7 $\pm$ 0.08} \textcolor{grey}{(94.2 $\pm$ 0.17)} & \underline{54.72 $\pm$ 0.48} \textcolor{grey}{(75.97 $\pm$ 0.35)} \\
Mixup + VarMixup & 83.92 $\pm$ 0.10 \textcolor{grey}{(96.78 $\pm$ 0.36)} & \textbf{58.22 $\pm$ 0.25} \textcolor{grey}{(79.3 $\pm$ 0.27)}\\
\bottomrule
\end{tabular}}
\end{center}
\end{table} 

\section{Conclusions}
\label{sec_conclusions}
\vspace{-5pt}
In this work, we proposed a Mixup-based vicinal distribution, VarMixup, which performs linear interpolation on an unfolded latent manifold where linearity in between training examples is likely to be preserved by construction. We show that VarMixup trained models are more robust to common input corruptions, are better calibrated and have significantly lower local-linear loss than regular Mixup models. As expected and noted earlier, in some places, we do observe a trade-off between clean and robust accuracy, and leave this as a direction for future works to explore. Additionally, our experiments indicate that VarMixup adds more off-manifold images to training than regular mixup, which we hypothesize is a key reason for the observed robustness. Our work highlights the efficacy of defining vicinal distributions by using neighbors on unfolded latent manifold rather than data manifold and we believe that our work can open a discussion around this notion of robustness and choice of vicinal distributions on generative latent spaces.


\section*{Acknowledgements}
\vspace{-5pt}
This work has been partly supported by the funding received from DST, Govt of India, through the IMPRINT program (IMP/2019/000250). We also acknowledge IIT-Hyderabad and JICA for provision of GPU servers for the work.

\bibliographystyle{model2-names}
\bibliography{refs}

\end{document}